\documentclass[runningheads]{llncs}
\usepackage{graphicx}
\usepackage{times}
\usepackage{epsfig}
\usepackage{graphicx}
\usepackage{amsmath}
\usepackage{amssymb}
\usepackage{pifont}
\usepackage{float}
\usepackage{xcolor}
\usepackage{bm}
\usepackage{times}
\usepackage{epsfig}
\usepackage{graphicx}
\usepackage{float}
\floatstyle{plaintop}
\restylefloat{table}
\usepackage{multirow}

\usepackage{array}
\newcolumntype{x}[1]{>{\centering\arraybackslash\hspace{0pt}}p{#1}}

\usepackage[skip=0.2\baselineskip]{caption}

\begin{document}

\title{Image Alignment in Unseen Domains \\ via Domain Deep Generalization}

\author{Thanh-Dat Truong\inst{1}, Khoa Luu\inst{2}, Chi Nhan Duong\inst{3}, Ngan Le\inst{4}, Minh-Triet Tran\inst{1}}

\institute{University of Science, VNU-HCM, Vietnam \\ \email{ttdat@selab.hcmus.edu.vn, tmtriet@fit.hcmus.edu.vn} \and
  			   University of Arkansas, USA \\ \email{khoaluu@uark.edu} \and 
			   Concordia University, Canada \\ \email{dcnhan@ieee.org} \and
			   Carnegie Mellon University, USA \\ \email{thihoanl@andrew.cmu.edu}
}

\maketitle 

\begin{abstract}

Image alignment across domains has recently become one of the realistic and popular topics in the research community. In this problem, a deep learning-based image alignment method is usually trained on an available large-scale database. During the testing steps, this trained model is deployed on unseen images collected under different camera conditions and modalities. The delivered deep network models are unable to be updated, adapted or fine-tuned in these scenarios. Thus, recent deep learning techniques, e.g. domain adaptation, feature transferring, and fine-tuning, are unable to be deployed.
This paper presents a novel deep learning based approach\footnote{The implementation of this work will be publicly available.} to tackle the problem of across unseen modalities. The proposed network is then applied to image alignment as an illustration.
The proposed approach is designed as an end-to-end deep convolutional neural network to optimize the deep models to improve the performance. 
The proposed network has been evaluated in digit recognition when the model is trained on MNIST and then tested on unseen MNIST-M dataset.
Finally, the proposed method is benchmarked in image alignment problem when training on RGB images and testing on Depth and X-Ray images.
\end{abstract}

\section{Introduction} \label{sec:intro}

\begin{figure}
	\centering \includegraphics[width=0.85\columnwidth]{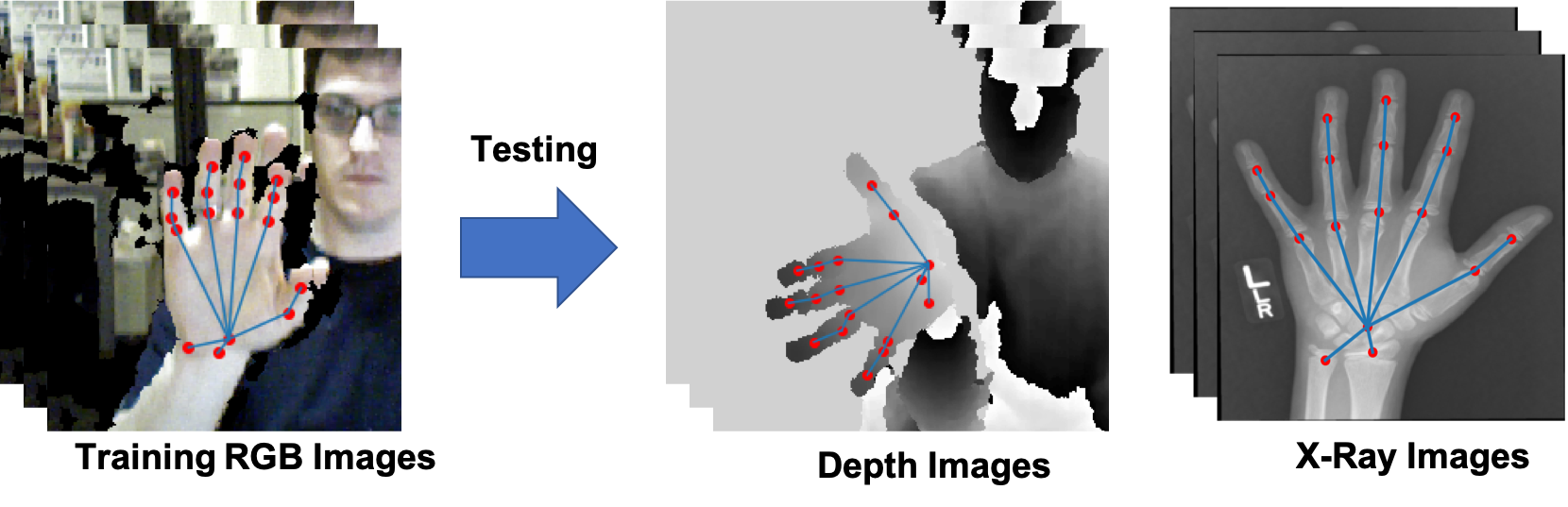}
	\caption{The Proposed Image Alignment Method across Image Modalities.}
	\label{fig:Fig1}
\end{figure}

Image alignment has an important role in numerous medical applications, e.g clinical track of events, clinical diagnostics, surgical procedure tracking, etc. Image alignment of medical images can be considered as a solved problem in the context of both training and testing images are collected under the same conditions. However, in numerous medical applications, due to the privacy, the deployment context or the limitations of the image collecting process, testing images are unavailable during the training steps.
Indeed, there is a need to develop a class of image alignment methods that only requires publicly available images, e.g. RGB images, in training, and then generalizes them in unseen domains, e.g. depth images or X-ray images as shown in Figure \ref{fig:Fig1}.

In order to approach the problem of image alignment across domains, domain adaptation \cite{pmlr-v37-ganin15} \cite{adda_cvpr2017} has recently received significant attention. In this method, the knowledge from the source domain will be learned and adapted to target domains. 
During the testing time, the trained model will be tested \textit{only} in the target domains. 
However, domain adaptation requires data in target domains in the training process. In many real-world applications, it is impossible to observe training samples from new unseen domains in the training process.
In such scenarios, domain adaptation is unable to be re-trained or fine-tuned with the inputs in new unseen domains.
Therefore, domain adaptation cannot be applied in these problems since the new unseen target domains are unavailable.

\subsection{Contributions of this Work}

In this paper, we present a novel \textbf{D}eep learning based \textbf{G}eneralized \textbf{I}mage \textbf{A}lignment (DeGIA) approach to image alignment across domains that can be applied in various medical applications. The proposed DeGIA requires only single source domain for training. It is then able to well generalize testing images in new unseen domains for the task of image alignment.
The proposed approach is designed in an end-to-end fashion and easily integrated with any CNN-based deep network deployed for medical image alignment. Through the experiments on RGB, Depth and X-Ray images, our proposal consistently helps to improve performance of medical image alignment.

To the best of our knowledge, this is the first time an image alignment approach across domains is introduced and achieves the improvement results.
The proposed image alignment method is also potentially applied in many other medical applications.

\section{Related Work} \label{sec:related_work}
% ...
\textbf{Medical Image Alignment} Image alignment for medical images has been early developed. They are feature-based methods, e.g. SIFT \cite{sift_ref}, SURF \cite{surf_ref},  popularly adopted for general computer vision applications, especially in image alignment.
Xiahai et. al. \cite{norm_vector_image_registration} presented a conception-normal vector information to evaluate the similarity between two images. 
Jingfan et. al. \cite{adversarial_similarity_image_aligment} introduced an unsupervised image alignment for medical images. The registration network is trained based on feedbacks from the discrimination network designed to judge whether a pair of registered images is aligned.

\noindent \textbf{Domain Adaptation} has recently become one of the most popular research topics in computer vision and machine learning \cite{pmlr-v37-ganin15} \cite{DBLP:journals/corr/TzengHDS15} \cite{Sener:2016:LTR:3157096.3157333} \cite{DBLP:journals/corr/TzengHZSD14} \cite{adda_cvpr2017}. Tzeng et al. \cite{adda_cvpr2017} proposed a framework for Unsupervised domain adaptation based on adversarial learning objectives.
Liu et al. \cite{NIPS2016_6544} presented Coupled Generative Adversarial Network to learn a joint distribution of multi-domain images.
Ganin et al. \cite{pmlr-v37-ganin15} proposed to incorporate both classification and domain adaptation to a unified network so that both tasks can be learned together.

\noindent \textbf{Domain Generalization}
Domain Generalization aims to learn an adversarially robust model that can well generalize in the most testing scenarios.
M. Ghifary et al. \cite{domain_generalization_Ghifary_2015_ICCV} proposed Multi-Task Autoencoder that learns to transform the original image into analogs in multiple related domains and makes features more robust to variations across domains.
Meanwhile, MMD-AAE \cite{domain_generalization_Li_2018_CVPR} tries to learn a feature representation by jointly optimizing a multi-domain autoencoder regularized via the Maximum Mean Discrepancy (MMD) distance. 
K. Muandet et al.\cite{domain_generalization_Muandet_ICML_2018} proposed a kernel-based algorithm for minimizing the differences in the marginal distributions of multiple domains.
Y. Li et al. \cite{domain_generalization_Li_2018_ECCV} presented an end-to-end conditional invariant deep domain generalization approach to leverage deep neural networks for domain-invariant representation learning.
R. Volpi et al. introduced Adversarial Data Augmentation (ADA) \cite{generalize-unseen-domain} to generalize to unseen domains. 

\section{The Proposed Method} \label{sec:proposed_method}
\begin{figure}[!t]
	\centering \includegraphics[width=0.8\columnwidth]{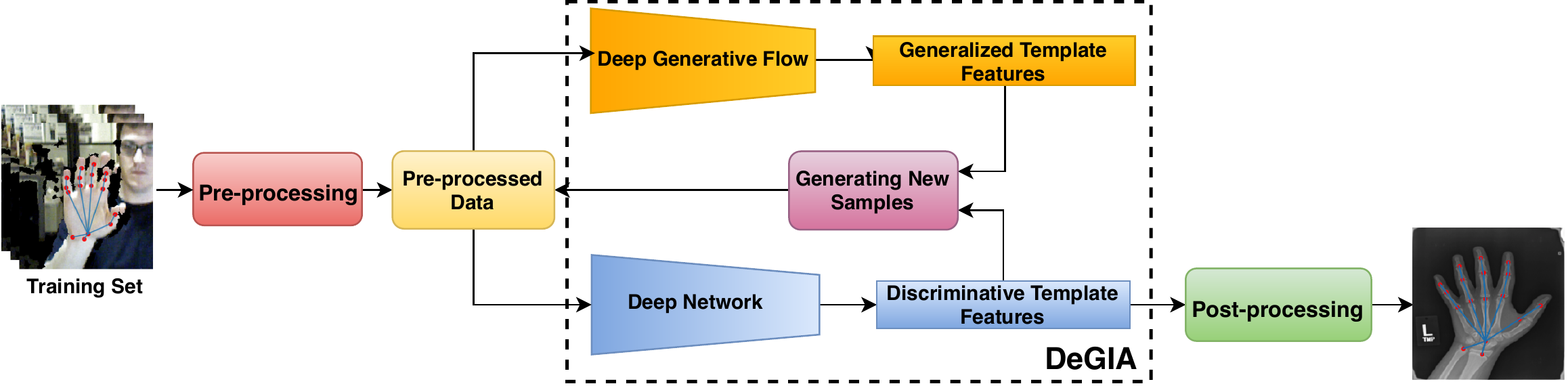}
	\caption{The proposed Deep learning based Generalized Image Alignment (DeGIA) method}
	\label{fig:Method}
\end{figure}

In this section, we introduce a novel \textit{Deep learning based Generalized Image Alignment} (DeGIA) approach that is able to simultaneously learn the discriminative template features for landmark points and generalize these template features so that they can represent the features in different domains.

The proposed deep network architecture is designed in an end-to-end fashion as shown in Figure \ref{fig:Method} with two main components: (1) Discriminative Template Feature Extraction with CNN architecture; and (2) Deep Generative Flows for generalizing these template features.
While the former is to generate template features of landmark points as much discriminative as possible to maximize the alignment precision, the latter aims to maintain the consistent structures of these features against the domain shifting to enhance the robustness of these features. Both of them are simultaneously optimized in DeGIA to generalize images in new unseen domains.

As a result, without seeing any data from other target domains, our learned model is still able to efficiently extract the \textit{generalized template features} for all landmark points and robustly detect them in different domains with high accuracy. The data in source domain can be considered as RGB hand images that are easy to collect, while the target domains consist of depth hand pose and X-ray images.

\subsection{Discriminative Template Feature Extraction}

Mathematically, image alignment can be formulated as a regression problem. In particular, the model receives an input image and then a regression function is built to detect $N_{\textit{lp}}$ landmark points. Let $\mathbf{X, Y}$ are input images and the corresponding ground-truths, $\mathcal{M}$ is a model, $\theta$ is a set of parameters, and $\mathbf{\bar{Y}}$ is detected landmark points. The image alignment problem can be formulated as $\mathbf{\bar{Y}} = \mathcal{M}(\mathbf{X};\theta)$. To learn the model, we use mean squared error (MSE) as the objective learning function of the model as in Eqn. (\ref{eqn:mse}).

\begin{equation}
    \small
     \ell_{\text{\textit{MSE}}}(\mathbf{X,Y};\mathcal{M})= || \mathbf{\bar{Y}} - \mathbf{Y} ||^2_2  = ||\mathcal{M}(\mathbf{X};\theta) - \mathbf{Y} ||^2_2
     \label{eqn:mse}
\end{equation}

\noindent The model $\mathcal{M}$ generally can be any deep network structure, e.g LeNet, AlexNet, ResNet, etc. The model learns to extract Discriminative Template Features that can be generalized to new unseen domains via the generative component in the DeGIA framework.

\subsection{Generalized Template Features} \label{sec:unseen_domain_gen}
Since pixel values in the image space are quite sensitive to the domain shifting, rather than directly learning and generalizing features in the original image space, we propose to employ these processes in a learned latent space where each data point is represented as density Gaussian distributions.
With these distributions, the structure of features can be efficiently maintained even when shifted among different feature domains.
By this way, the learned features can be robustly generalized. The mapping process from the image space to the latent space is named as the deep generative flow.

In particular, let $P_Z^{src} = \large{\bigcup_c} \text{ }  \mathcal{N}(\boldsymbol{\mu}^{src}_c, \Sigma^{src}_c)$ be $C$ Gaussian distributions with different means $\{\boldsymbol{\mu}^{src}_1, \boldsymbol{\mu}^{src}_2, .., \boldsymbol{\mu}^{src}_C\}$ and covariances $\{\Sigma^{src}_1,\Sigma^{src}_2,...,\Sigma^{src}_C\}$ of $C$ classes.
Then a generative flow is formulated as learning a function $\mathcal{F}$ that maps an image $\mathbf{x}$ in the image space of the source domain to its latent representation $\mathbf{z}$ 
in the latent space $\mathcal{Z}$ such that the density function $P_X^{src}(\mathbf{x})$ can be indirectly estimated via $P_Z^{src}(\mathbf{z})$:

\begin{equation} \label{eqn:DensityFunc}
\small
    p_X^{src}(\mathbf{x},c;\theta_1) = p_Z^{src}(\mathbf{z},c;\theta_1)\left|\frac{\partial \mathcal{F} (\mathbf{z},c;\theta_1)}{\partial\mathbf{x}} \right|
\end{equation}
where $p_X(\mathbf{x},c)$ defines the density distributions of a sample of class label $c \in \mathcal{C}$ ($\mathcal{C}$ is the number of classes) in the image space; similarity, $p_Z(\mathbf{z},c;\theta_1)$ defines the density distributions of a sample of class label $c$ in the latent space.
$\frac{\partial \mathcal{F} (\mathbf{z}, c;\theta_1)}{\partial\mathbf{x}}$ denotes the Jacobian matrix with respect to $\mathbf{x}$. We adopt the structure of \cite{Duong_2017_ICCV} for $\mathcal{F}$ and learn its parameters by maximizing the log-likelihood as in Eqn. \eqref{eqn:DensityFunc}.

\begin{equation}
\small
    \theta_1^* = \arg \max_{\theta_1} \sum_{i}\log p_X(\mathbf{x}^i,c;\theta_1)
\end{equation}

Then, let $\rho$ be the distance between a new unseen distribution $P_X$ and the distribution of source domain $P_X^{src}$ in latent space. The goal of generalization process is to learn the Generalized Template Features such that even when the new unseen distribution $P_X$ is deviated from $P_X^{src}$ a distance $\rho$, these learned features for landmarks can still minimize the alignment loss $\ell(\mathbf{X,Y};\mathcal{M}, \theta)$ as follows,

\begin{equation} \label{Eqn:WorseCaseFormulation}
\small
    \arg \min_{\theta} \sup_{P:d(P_X,P_X^{src})\leq \rho} \mathbb{E}\left[ \ell(\mathbf{X,Y};\mathcal{M}, \theta)\right] 
\end{equation}
In the next section, our DeGIA is presented to solved Eqn. \eqref{Eqn:WorseCaseFormulation} for generalization.

\subsection{Deep Learning Based Generalized Image Alignment} \label{subsec:degia}

The proposed DeGIA consists of two main components: (1) \textit{Discriminative Template Feature} and (2) \textit{Generalized Template Features}. 
Firstly, given a training dataset, parameters $\{\theta_1, \theta\}$ of $\mathcal{F}$ and the network $\mathcal{M}$ are updated according to the loss function as follows,

\begin{equation} \label{eqn:LossFunction_All}
\small
    \ell(\mathbf{X,Y},\mathcal{C};\mathcal{M},\mathcal{F},\theta, \theta_1) = \ell_{\text{\textit{MSE}}}(\mathcal{M}(\mathbf{X};\theta),\mathbf{Y}) - \log p_X(\mathbf{X},\mathbf{C}; \theta_1)
\end{equation}
where the first term is the specific task loss for $\mathcal{M}$ and the second term is the log-likelihood of $\mathcal{F}$. 
Notice that during the optimization process in DeGIA, when discriminative features are updated, the generative flow also requires to be updated accordingly. Therefore, both alignment and log-likelihood losses are included in the objective function as in Eqn. \eqref{eqn:LossFunction_All}.
\noindent Secondly,
applying the Lagrangian relaxation to Eqn. \eqref{Eqn:WorseCaseFormulation}:
\begin{equation} \label{eqn:WorseCaseFormulation}
\small
\arg \min_{\theta_1} \sup_{P_X} \mathbb{E} 
\left[\ell(\mathbf{X,Y},\mathcal{C};\mathcal{M},\mathcal{F}, \theta, \theta_1)\right] -
\alpha \cdot \Big[d(P_X,P_X^{src}) + ||\mathcal{M}(\mathbf{X}) - \mathcal{M}(\mathbf{X}^{src}_X)||^2_2 \Big])
\end{equation}
where $d(\cdot, \cdot)$ is the distance between probability distributions; $P^{src}_X$ and $P_X$ are the density distributions of the source and current expanded domains, respectively. 

In order to learn the model that optimizes Eqn. \eqref{eqn:WorseCaseFormulation}, we employ a learning process with two phases: (1) Minimization phase to optimize $\mathcal{M}$ and $\mathcal{F}$; and (2) Maximization phase to approximate $P_X$. For the minimization phase, we use samples from training set to minimize the objective $\ell(\mathbf{X,Y},\mathcal{C};\mathcal{M},\mathcal{F},\theta, \theta_1)$. Meanwhile, the maximization phase generates new samples of $P_X$ by maximizing Eqn. \eqref{eqn:max_argsol}.
\begin{equation} \label{eqn:max_argsol}
\small
\mathbf{x}=\arg \max_{\mathbf{x}} \{ \ell(\mathbf{x,y},c;\mathcal{M},\mathcal{F}, \theta, \theta_1) - \alpha \cdot \Big[d(P_x,P_x^{src}) + ||\mathcal{M}(\mathbf{x}) - \mathcal{M}(\mathbf{x}^{src}_x)||^2_2 \Big] \}
\end{equation}
In this way, new generated samples not only well generalize to unseen domains but also ``hard" samples under the current models due to maximizing loss of the network. Then, new generated samples are added to the training set to update $\mathcal{M}$ and $\mathcal{F}$ in the next training steps. This process continues until convergence.

\noindent In order to better generalize, instead of pre-defining mean and corvariances of the deep generative flow, we design learnable functions $\mathcal{G}_m(c)$ and $\mathcal{G}_{std}(c)$ that map given label $c$ to corresponding mean and corvariance. To make more robust, we add a noise signal $\mathbf{n}$, specifically,
\begin{equation} \label{eqn:mean_sig}
\small
\begin{split}
    \boldsymbol{\mu}_c & = \gamma \mathcal{G}_m(c) + \lambda \mathcal{H}_m(\mathbf{n})\\
    \mathbf{\Sigma}_c & = \mathcal{G}_{std}(c)
\end{split}
\end{equation}
where $\mathcal{H}_m(\mathbf{n})$ is a flexible shifting range, $\gamma$ and $\lambda$ are hyper-parameters that control the separation of the Gaussian distributions between different classes and the contribution of $\mathcal{H}_m(\mathbf{n})$ to $\boldsymbol{\mu}_c$.

\section{Experiments} \label{sec:experiments}

This section presents experimental results of DeGIA on benchmark datasets. 
Firstly, we shows the effectiveness of our proposed methods with ablative experiments in Sec. \ref{sec:ablation_study}.
In these experiments, MNIST is used as the only single source training set and MNIST-M plays a role as an unseen test set. 
As the achieved results, our proposals help consistently to improve performance of deep networks. 
Finally, we do evaluate DeGIA on hand pose alignment task trained on RGB NYU Hand Pose images \cite{nyu_data} and tested Depth NYU Hand Pose images \cite{nyu_data} and X-Ray images \cite{x_ray_data}. 

\begin{figure}[!b]
	\centering \includegraphics[width=0.99\columnwidth]{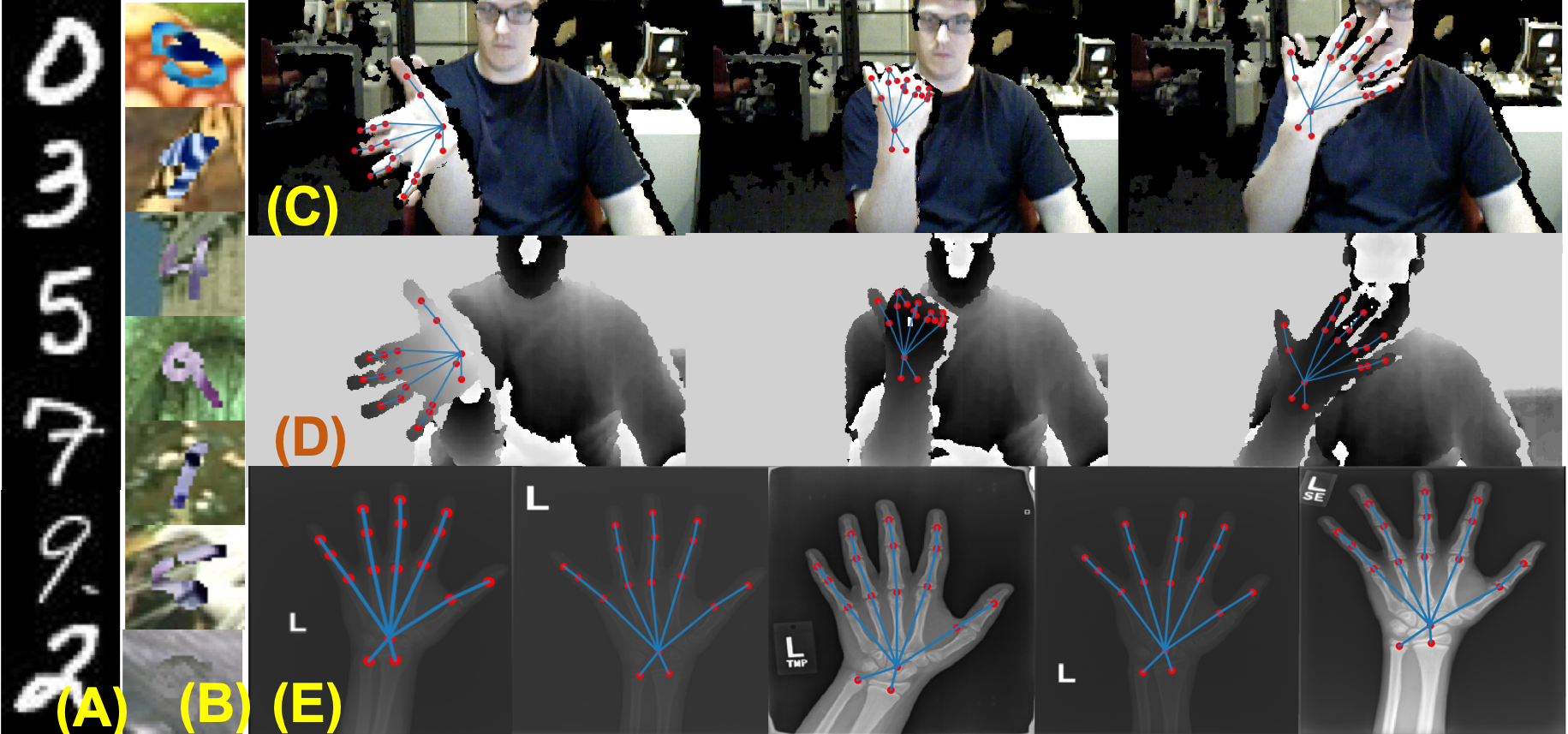}
	\caption{Some example images used in our experiments: (A) MNIST Dataset, (B) MNIST-M Dataset, (C) NYU RGB Hand Poses, (D) NYU Depth Hand Poses and (E) X-Ray Images.}
	\label{fig:hand_pose_data}
\end{figure}

\subsection{Ablation Study} \label{sec:ablation_study}

\begin{table}[!h]
    \centering
    \caption{Ablative experiment results (\%) on effectiveness of the parameters $\lambda$, $\alpha$
    and $\beta$ that control the distribution separation and shitting range. MNIST is used as the only training set, MNIST-M is used as the unseen testing set.}
    \label{tab:ablation}
\begin{tabular}{|x{1.5cm}|x{1.5cm}|x{0.75cm}|x{0.75cm}|x{0.75cm}||x{0.75cm}|x{0.75cm}|x{0.75cm}||x{0.75cm}|x{0.75cm}|x{0.75cm}|x{0.75cm}|}
    \hline
     \multirow{2}{*}{\textbf{Dataset}}& \multirow{2}{*}{\textbf{Methods}} & \multicolumn{3}{c||}{\boldsymbol{$\lambda$}} & \multicolumn{3}{c||}{\boldsymbol{$\alpha$}} & \multicolumn{4}{c|}{\boldsymbol{$\beta (\%)$}} \\
     \cline{3-12}
     \multirow{2}{*}{}& \multirow{2}{*}{} & $\textbf{0.0}$ & $\textbf{0.1}$ & $\textbf{1.0}$ & $\textbf{0.01}$ & $\textbf{0.1}$ & $\textbf{1.0}$ & $\textbf{1\%}$ & $\textbf{10\%}$ & $\textbf{20\%}$ & $\textbf{30\%}$ \\
    \hline
    
     \multirow{2}{*}{MNIST}     & {Pure-CNN}        & \multicolumn{10}{c|}{99.06} \\ 
     \cline{2-12}

     \multirow{3}{*}{}          & \textbf{DeGIA}   & \textbf{99.43} & 99.06 & 99.24 & 99.40 & \textbf{99.39} & 99.06 & 99.40 & 99.36 & 99.06 & \textbf{99.42} \\
     \hline
     
     \multirow{2}{*}{MNIST-M}   & {Pure-CNN}        & \multicolumn{10}{c|}{56.47} \\ 
     \cline{2-12}
          
     \multirow{3}{*}{}          & \textbf{DeGIA}   & 57.53 & \textbf{62.94} & 56.69 & 60.13 & 57.62 & \textbf{62.94} & 62.32 & 60.54 & \textbf{62.94} & 58.52 \\
     \hline
\end{tabular}
\end{table}

First of all, we evaluate our proposed methods on MNIST (Fig. \ref{fig:hand_pose_data}(A))  and MNIST-M (Fig. \ref{fig:hand_pose_data}(B)). This experiment aims to measure the effectiveness of hyper-parameters to our proposed method. The training process just takes MNIST as a source training domain. Meanwhile, MNIST-M will be a testing domain (an unseen domain). We choose LeNet \cite{lenet_ref} as deep network classifier, learning rate and batch size are set to $0.0001$ and $128$, respectively.

As we mentioned in Sec. \ref{subsec:degia}, there are two alternating phases in the training process: (1) Minimization phase optimizes the $\mathcal{M}$ and $\mathcal{F}$ and (2) Maximization (perturb) phase approximates $P_X$. For the maximization phase, we randomly select $\beta$ percent of the number of training images to explore new samples in unseen domains. We consider the effect of percentage ($\beta$) of new generated samples, specifically, we do evaluate $\beta \in \{1\%, 10\%, 20\%, 30\%\} $.
Generalized template feature extraction is handled by a set of scale parameters, i.e $\gamma$ and $\lambda$, are control numbers of distribution separation and shitting range as shown in Eqn. \eqref{eqn:mean_sig}. As the results shown in Table \ref{tab:ablation}, our proposed approach helps to improve the classifiers significantly, specifically, the accuracy of MNIST-M has been improved \textbf{6.47\%}. Since the testing phase just uses the discriminative template feature extraction branch, the inference time of DeGIA will be the same with the stand-alone CNN.

\subsection{Deep learning based Generalized Hand Pose Alignment}

In this experiment, we aim for improving Hand Pose Alignment trained on NYU Hand Pose RGB images (Fig. \ref{fig:hand_pose_data}(C)) \cite{nyu_data} and tested on NYU Hand Pose Depth (Fig. \ref{fig:hand_pose_data}(D)) and X-Ray images  (Fig. \ref{fig:hand_pose_data}(E)) \cite{x_ray_data}. 
In our experiment, we select $N_{\textit{lp}} = 17$ of a hand as landmark points.
We use features of the deep network (i.e LeNet, AlexNet, VGG, ResNet, DenseNet) forward to a fully connected layer to detect $N_{\textit{lp}} = 17$ landmark points, all images were resized to $256 \times 256$. 
In the maximization phase, we randomly select 500 images to generate new samples.
We compare our method DeGIA against stand-alone network (Pure-CNN). As the results in Table \ref{tab:hand_alignment}, our DeGIA robustly detects landmark points in new unseen domains (i.e depth and X-Ray images) and achieves state-of-the-art results.

\begin{table}[!h]
    \centering
    \caption{Experimental Results (MSE) on Hand Alignment with various common deep network structures.}
    \label{tab:hand_alignment}
    
    \begin{tabular}{| x{2cm} | x{2cm} | x{2cm} | x{2cm} | x{2cm} | x{2cm} |}
        \hline
        \textbf{Networks} & \textbf{Methods} & \textbf{RGB} & \textbf{Depth} & \textbf{X-Ray} \\
        \hline
        \multirow{2}{*}{LeNet}      & Pure-CNN          &  0.0506  &  0.0697  &  0.1266 \\
        \cline{2-5}
        \multirow{2}{*}{}           &  \textbf{DeGIA}  &  \textbf{0.0345}  &  \textbf{0.0534}  &  \textbf{0.0735} \\
        \hline
        \multirow{2}{*}{AlexNet}    & Pure CNN          &  0.0105  &  0.0484  &  0.0472 \\
        \cline{2-5}
        \multirow{2}{*}{}           &  \textbf{DeGIA}  &  \textbf{0.0102}  &  \textbf{0.0401}  &  \textbf{0.0336} \\
        \hline
        \multirow{2}{*}{VGG}        & Pure CNN          &  \textbf{0.0083}  &  0.0822  &  0.0981 \\
        \cline{2-5}
        \multirow{2}{*}{}           &  \textbf{DeGIA}  &  0.0195  &  \textbf{0.0709}  &  \textbf{0.0284} \\
        \hline
        \multirow{2}{*}{ResNet}     & Pure CNN          &  0.0982  &  0.2596  &  0.1447 \\
        \cline{2-5}
        \multirow{2}{*}{}           &  \textbf{DeGIA}  &  \textbf{0.0295}  &  \textbf{0.1065}  &  \textbf{0.1140} \\
        \hline
        \multirow{2}{*}{DenseNet}   & Pure CNN          &  0.0194  &  0.0958  &  0.1221 \\
        \cline{2-5}
        \multirow{2}{*}{}           &  \textbf{DeGIA}  &  \textbf{0.0162}  &  \textbf{0.0863}  &  \textbf{0.1214} \\
        \hline
    \end{tabular}
    \vspace{-7mm}
\end{table}

\section{Conclusions}
\label{sec:concl}

This paper has presented a novel DeGIA approach to medical image alignment across domains. 
The proposed method requires only a single source domain for training and well generalizes in unseen domains.
It is designed within an end-to-end CNN to extract discriminative template features and generalized template features to robustly detect landmarks.
It has been benchmarked on numerous databases across domains, including MNIST, MNIST-M, NYU RGB and Depth hand pose images, and X-Ray hand images, and consistently achieved improvement performance results. The proposed image alignment method is also potentially applied in many other medical applications.
\bibliographystyle{splncs04}
\small 
\bibliography{references}
\end{document}